\documentclass{svproc}

\usepackage{url}
\usepackage{subfig}
\usepackage{graphicx}
\usepackage{colortbl}
\usepackage{PTSansNarrow}
\usepackage[T1]{fontenc}
\usepackage{array,tabularx}
\usepackage[most]{tcolorbox}
\usepackage{xcolor}
\usepackage[numbers]{natbib}
\usepackage{authblk} 
\usepackage{dsfont} 
\usepackage{amsmath}
\usepackage{soul}
\usepackage{fourier}
\DeclareMathAlphabet{\mathcal}{OMS}{cmsy}{m}{n}
\SetMathAlphabet{\mathcal}{bold}{OMS}{cmsy}{b}{n}
\newcommand{\bigO}{\mathcal{O}}

\newcommand{\er}{Erdős–Rényi }

\newcommand{\ONE}{\mathds{1}}
\newcommand{\DEF}{\overset{\textbf{def}}{=}}

\title{LLMs Prompted for Graphs: Hallucinations and Generative Capabilities}

\author{Gurvan Richardeau\inst{1,2}, Samy Chali\inst{1}, Erwan Le Merrer\inst{2}, Camilla Penzo\inst{1}, Gilles Tredan\inst{3}}

\institute{
PEReN, France \and
Univ Rennes, Inria, CNRS, IRISA, France \and
LAAS/CNRS, France
}
\begin{document}

\maketitle

\begin{abstract}
Large Language Models (LLMs) are nowadays prompted for a wide variety of tasks. In this article, we investigate their ability in reciting and generating graphs.
We first study the ability of LLMs to regurgitate  well known graphs from the literature (e.g. Karate club or the graph atlas)\footnote{This article is an extended version of the conference paper \cite{merrer2024llms}.}.
Secondly, we question the generative capabilities of LLMs by asking for \er random graphs. As opposed to the possibility that they could memorize some \er graphs included in their scraped training set, this second investigation aims at studying a possible emergent property of LLMs.
For both tasks, we propose a metric to assess their errors with the lens of hallucination (i.e. incorrect information returned as facts).
We most notably find that the amplitude of graph hallucinations can characterize the superiority of some LLMs. Indeed, for the recitation task, we observe that graph hallucinations correlate with the Hallucination Leaderboard, a hallucination rank that leverages $10,000$ times more prompts to obtain its ranking. For the generation task, we find surprisingly good and reproducible results in most of LLMs. We believe this to constitute a starting point for more in-depth studies of this emergent capability and a challenging benchmark for their improvements.
Altogether, these two aspects of LLMs capabilities bridge a gap between the network science and machine learning communities.

\keywords{Large language models, graphs, hallucinations, Erdős–Rényi, random generation.}
\end{abstract}

\section{Introduction}

Large Language Models (LLMs) recently attracted a lot of attention, thanks to sustained research efforts and a large spectrum of envisioned applications. This triggers a growing demand for tools to test and analyze these complex and expensive-to-set-up objects. In particular, methods to efficiently identify, differentiate, watermark LLMs, and methods to compare their accuracy and notably the potential presence of hallucinations are devised \cite{nadeau2024benchmarkingllama2mistralgemma,tonmoy2024comprehensivesurveyhallucinationmitigation,sansford2024grapheval}. The common denominator of these methods is the will to efficiently extract information from the LLM under scrutiny.

To achieve this information extraction, one can distinguish two broad categories of approaches: \emph{white box} approaches that rely on exploiting a privileged access to the model internals (e.g. probits, activation patterns, source code, model hyperparameters and weights), and \emph{black box} approaches that merely allow an auditor to interact with a given model. White box approaches provide precise answers, but are not always feasible, for instance, to an external auditor assessing a closed source LLM. In such cases, black box approaches are necessary and constitute the focus of this article.

To conduct a black box audit of a LLM, a common approach is to rely on multiple choice questions (MCQ) and test the accuracy of its answers. Using MCQ circumvents the difficulty and subjectivity of evaluating (a potentially large number of) answers to open questions. To this end, several  MCQ datasets are readily available, and cover a wide range of knowledge and reasoning domains. For instance, MedMCQA \cite{medmcqa} contains 182k four-choice questions assessing the target model general medical knowledge. Using this dataset yields 2 bits of information on the target model per requested prompt. This might reveal inefficient if the auditor has to pay for each prompt or if she has to limit the rate at which prompts are sent.

In this article, we explore a different approach using prompts that aims at obtaining \textit{graphs} rather than text. As graphs are structured, requesting graphs circumvents the problem of analyzing free form text answers. And most importantly, a single request to obtain a $n$ node graph from a LLM yields $\bigO(n^2)$ structured information bits from that model, enabling a much more efficient information extraction rate per request for the auditor. 
For instance, requesting the standard ``Les Misérables" graph is much more efficient than issuing queries like "do Jean Valljean and Cosette both appear in a chapter?", for any two of the 77 characters of the book.  
Our first research question, RQ1, focuses on studying the LLM ability to recite graphs they most likely have viewed during their training.
Indeed, throughout the years, network science has produced a number of standard benchmark graphs that are probably already ingested by most LLMs and that provide a readily available ground truth, e.g. in the SNAP \cite{snapnets} or KONECT \cite{konect} repositories among others), and as the training data of LLMs is known to include most of online scrapable data \cite{gao2020pile}.
While with RQ1 we question to which extent LLMs are able to memorize given examples of graphs, with our second research question RQ2, we interrogate a very different ability, generating randomness in a structured form. We choose \er random graphs. 
RQ2 is inspired by studies on randomness generations with LLMs \cite{hopkins2023can,vankoevering2024randomrandomevaluatingrandomness,harrison2024comparisonlargelanguagemodel} and on studies on LLMs emergent abilities. In the literature, an emergent ability can be defined as an \textit{ability to reason about novel problems in a zero-shot setting, i.e. without any training on the sought task} \cite{webb2023emergent}.

Though querying LLMs for graph structures is unexplored and might sound promising, it raises a number of questions we further investigate in the following.
First, we detail the interaction framework in which the objective is to collect a graph as a LLM output. Once this has been established, we focus on the quality of the collected graphs proposing adapted metrics to answer RQ1 and RQ2, within the lens of hallucination. Hallucination is commonly defined as when \textit{the generated content is nonsensical or unfaithful to the provided source content} \cite{ji2023survey, zhang2023siren}. In RQ1, as there is a ground truth comparison, any error can be interpreted as hallucination. In RQ2, we measure the frequency with which LLMs produce highly inaccurate random graphs, referring to such errors as hallucinations.

Finally, we showcase the potential of graph queries and their ability to extract information from LLMs in a request-efficient manner. To this end, we compare the Hallucination Leaderboard \cite{board} rank to the rank we produce with our two proposed metrics.

\section{The Topologies of LLM Graph Hallucinations (RQ1)}
\label{s:topologies}

Our aim is to ask LLMs for famous graphs for which the ground truth topology is commonly available online (see e.g. on repositories such as SNAP \cite{snapnets} or KONECT \cite{konect}), and consequently most likely part of the training set used for these LLMs \cite{gao2020pile}. We chose three graphs, the Zachary's Karate Club graph (coined KC hereafter) and Les Misérables (coined LM, see their description e.g. in the NetworkX library \cite{networkx}) and the 50th graph of the graph atlas \cite{read1998atlas}.

\paragraph{Prompting LLMs.} The prompts we used in order to obtain a graph structure are the following: \texttt{Provide me the so called ``X" graph as a python edge list; print it}, with X being the graph of interest, for instance the KC or the LM. We also used the prompt \texttt{Provide me with graph \# from the Graph Atlas, as a python edge list; print it}, with \texttt{\#} being the graph number in the case of queries concerning the graph atlas. We asked for Python structured responses because we use the NetworkX library \cite{networkx} to instantiate and analyze the outputs.

\paragraph{LLM Outputs.} Given the chosen prompts, the outputs are most often in the form of a list, e.g. $(1,2), (1,3), \dots$ (see Section \ref{appendix:output} for an example). The edge list is then parsed and built as a NetworkX graph; we used undirected graphs in alignment with the ground truth graphs chosen in this article. As the ground truth graphs are also present in the NetworkX library, the comparison is convenient.
To be noted that incomplete outputs (i.e. incomplete responses leading to partial edge lists) are nevertheless examined. We report that a small fraction of queried LLMs refused to return an edge list and claimed they do not have access to the data (as also noted by the Hallucination Leaderboard project \cite{board}). Some other LLMs only provide the Python code to print the queried graph using a certain library (e.g. NetworkX); these models are discarded from our study. 

\paragraph{Prompted LLMs.} The simplicity and low volume of necessary queries (prompts) enables a non-automated and full online experience (as opposed to downloading the model or using the API interface). We used online platforms that allow prompting access to LLMs via a web browser. In this work, we used the following platforms: Mistral \cite{mistral}, Vercel AI SDK \cite{vercel}, HuggingChat \cite{huggingchat}, ChatGPT \cite{chatgpt}, together.ai \cite{together} and Google's Gemini \cite{gemini}, with the default parameters the platforms defined for their hosted models.

\paragraph{Comparing output graphs to the ground truth.} In RQ1 we perform topological comparisons without considering the node labels returned by the LLM (e.g. characters names for the Les Misérables graph). These vary significantly, and thus a hard label matching would discard output graphs that might be topologically close. We leave a study on the labeling mismatch to future work.

\subsection{Statistics on the Topology of Output Graphs}

\begin{figure}[h!]
\centering
\subfloat[Output graph]{\label{a:graphs-a}\includegraphics[width=0.25\textwidth]{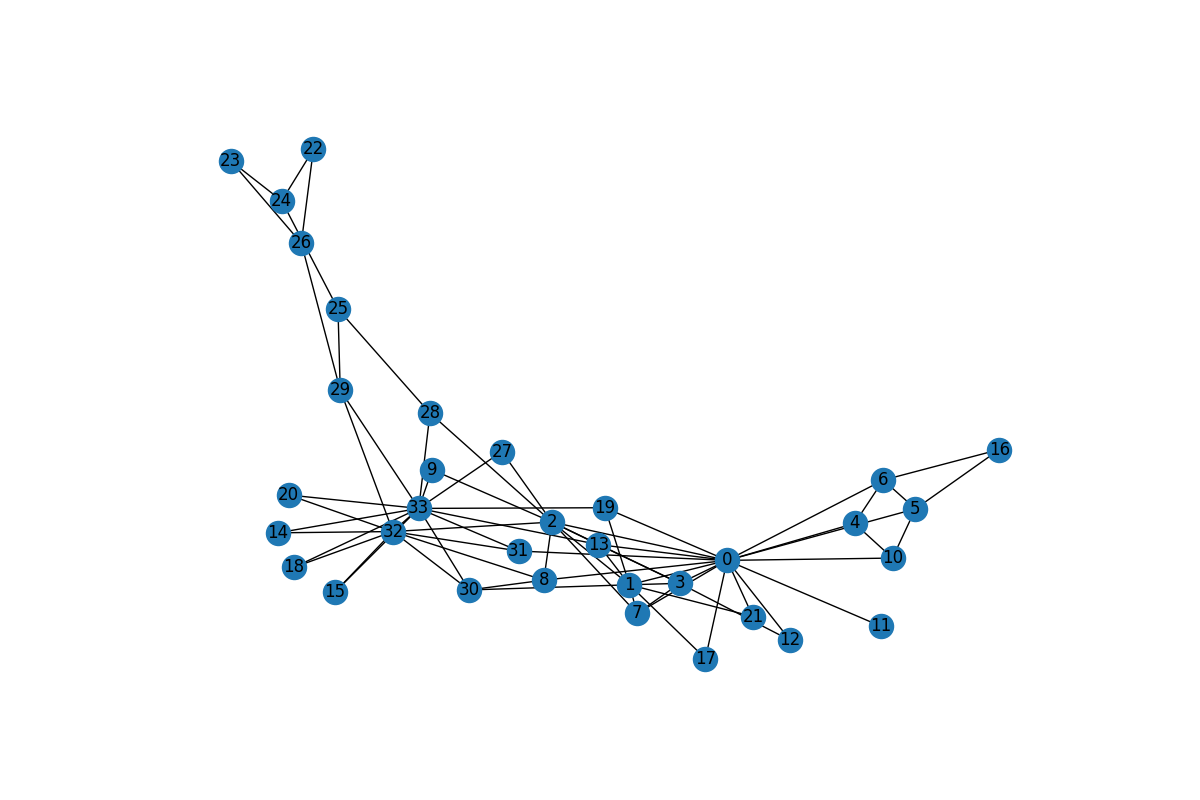}}
\hfill
\subfloat[Intersection]{\label{a:graphs-b}\includegraphics[width=0.25\textwidth]{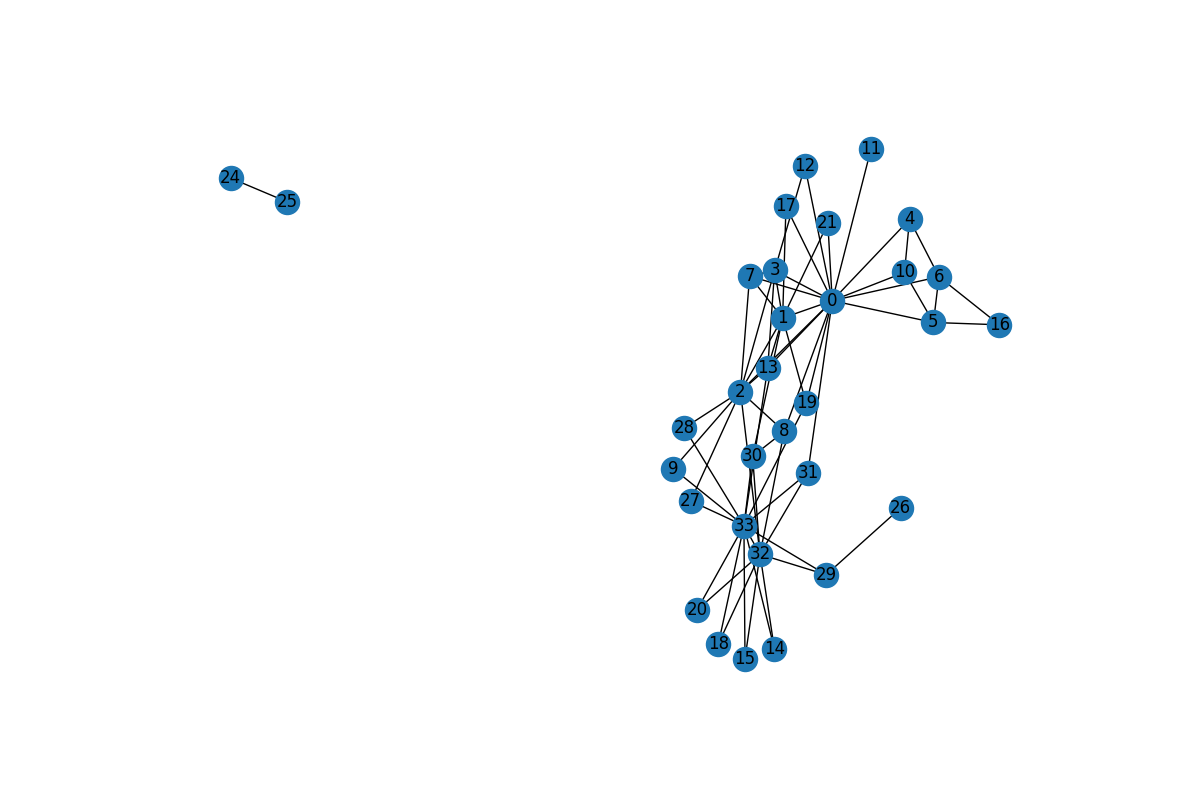}}%
\hfill
\subfloat[Added edges]{\label{a:graphs-c}\includegraphics[width=0.25\textwidth]{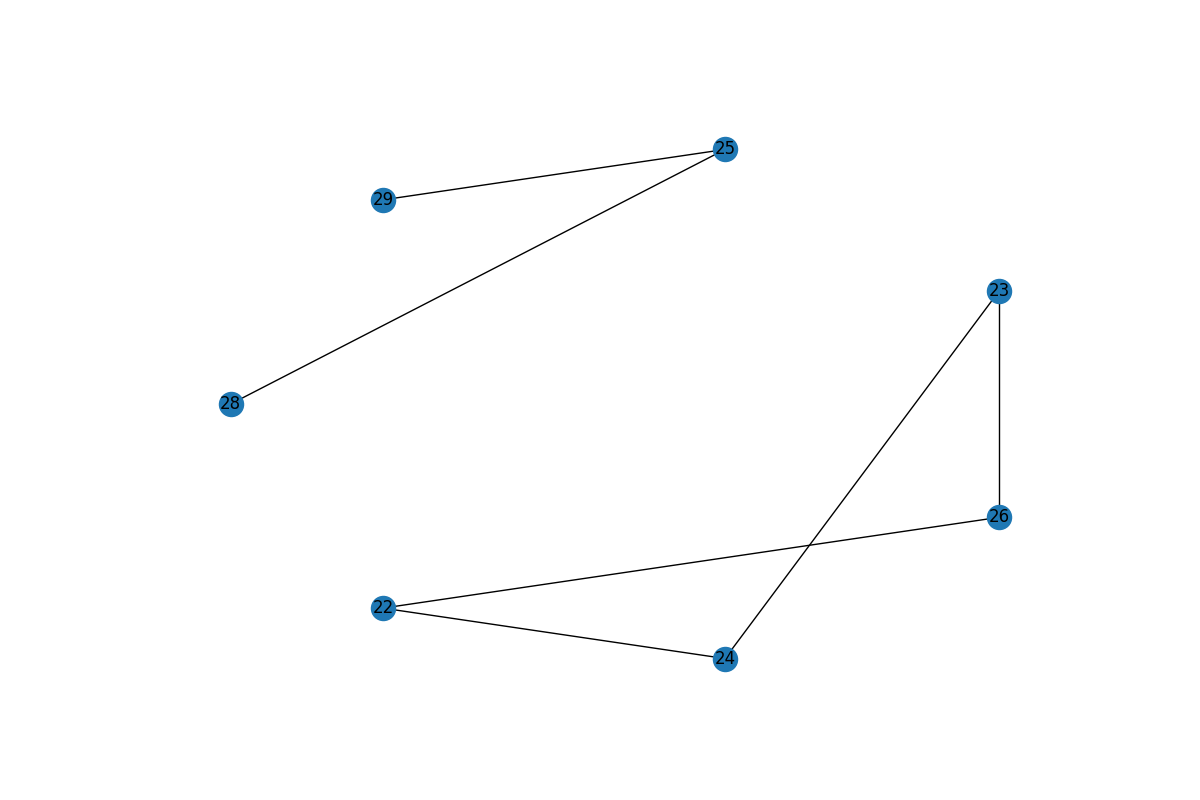}}%
\hfill
\subfloat[Missing edges]{\label{a:graphs-d}\includegraphics[width=0.25\textwidth]{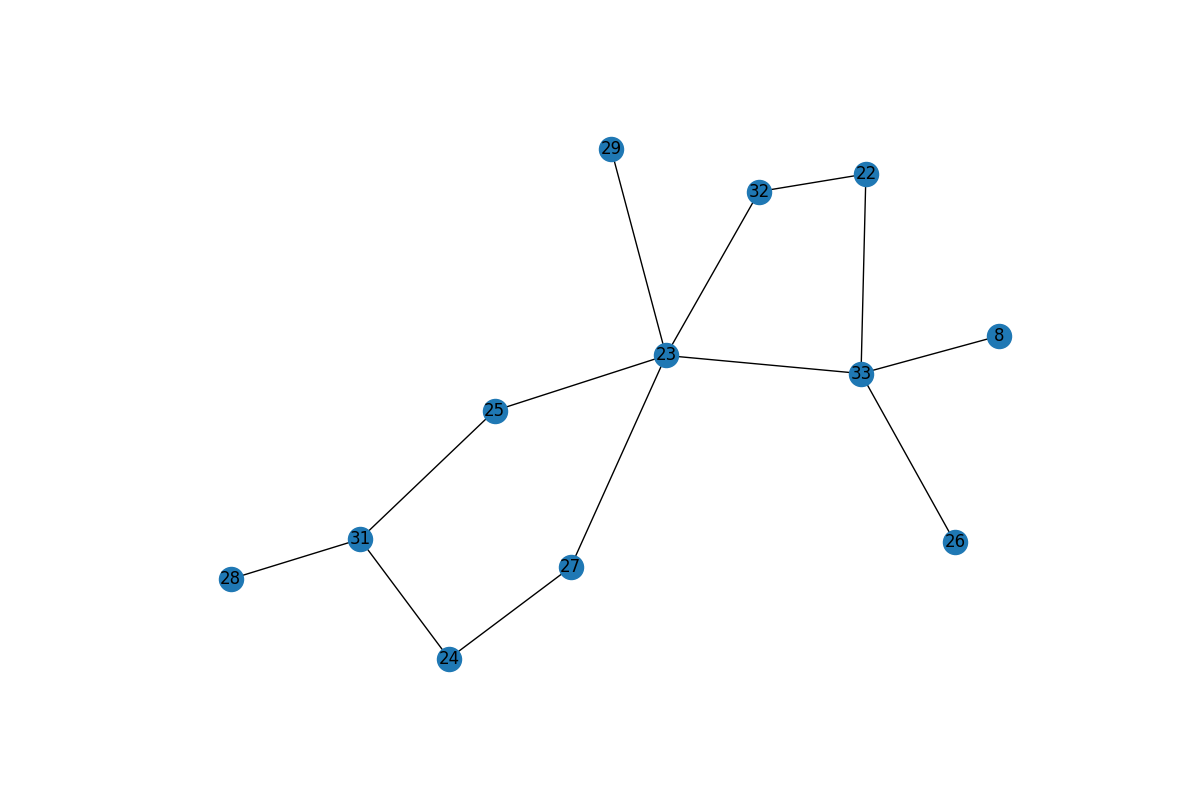}}%
\caption{Prompting gpt4o for the Zachary's karate club graph. (a) answered output graph, (b) intersection of the output graph with the KC ground truth graph, (c) the edges added (hallucinated) w.r.t. to the KC graph, and (d) the edges missing w.r.t. the KC graph.}
\label{graphs}
\end{figure}

We report in detail the prompting of 21 LLMs when we asked for a Zachary's Karate Club (KC) graph. Some other statistics on Les Misérables and graph atlas 50 are deferred in Appendix \ref{a:others}. 
As each prompt results in a graph, one can directly perform topological comparison between the ground truth and a LLM output graph. Such a comparison is represented in Figure \ref{graphs}. Figure \ref{a:graphs-a} shows the raw output graph, while Figure \ref{a:graphs-b} shows the graph intersection of the KC graph with the output graph from Figure \ref{a:graphs-a}. Figure \ref{a:graphs-c} presents the hallucinated edges (i.e. edges not present in KC but present in the output graph). Finally, Figure \ref{a:graphs-d} presents the edges that are forgotten in the output graph compared to the KC, when prompting gpt4o. We note that the result provided by gpt4o is relatively accurate in comparison to the others, as we can also see in Table \ref{table:stats}.

\begin{table}[b!]
\centering
\scriptsize{
\begin{tabular}{|c|cccccc}
\hline
\cellcolor[HTML]{DDDDDD}\textbf{LLM} & \multicolumn{1}{c|}{\cellcolor[HTML]{DDDDDD}\textbf{|V|}} & \multicolumn{1}{c|}{\cellcolor[HTML]{DDDDDD}\textbf{|E|}} &
\multicolumn{1}{c|}{\cellcolor[HTML]{DDDDDD}\textbf{density}} & \multicolumn{1}{c|}{\cellcolor[HTML]{DDDDDD}\textbf{assort.}} & \multicolumn{1}{c|}{\cellcolor[HTML]{DDDDDD}\textbf{modularity}}  & \multicolumn{1}{c|}{\cellcolor[HTML]{DDDDDD}\textbf{dist. to KC deg. seq.}} \\ \hline
(reference: KC ground truth)\cellcolor[HTML]{EEEEEE}  &  34 \cellcolor[HTML]{EEEEEE} &  78\cellcolor[HTML]{EEEEEE} &   0.14 \cellcolor[HTML]{EEEEEE} &  -0.48\cellcolor[HTML]{EEEEEE}  &  0.31 \cellcolor[HTML]{EEEEEE} &  0.0\cellcolor[HTML]{EEEEEE} \\ \cline{1-1}

dbrx-instruct  &  34  &  80  &  0.14  &  -0.47  &  0.15  &  2.0 \\ \cline{1-1}
gpt35  &  34  &  71  &  0.13  &  -0.41  &  0.36  &  3.74 \\ \cline{1-1}
gpt4o  &  34  &  71  &  0.13  &  -0.41  &  0.36  &  3.74 \\ \cline{1-1}
llama-3.1-70B-Instruct-Turbo  &  30  &  68  &  0.16  &  -0.29  &  0.4  &  8.6 \\ \cline{1-1}
gemini  &  16  &  21  &  0.17  &  -0.06  &  0.42  &  8.72 \\ \cline{1-1}
llama-3-8b-instruct  &  12  &  20  &  0.3  &  -0.0  &  0.29  &  10.0 \\ \cline{1-1}
llama-2-13b-chat-hf  &  6  &  8  &  0.53  &  -0.23  &  0.0  &  11.14 \\ \cline{1-1}
llama-3-sonar-small-32k-chat  &  13  &  31  &  0.4  &  -0.28  &  0.0  &  11.36 \\ \cline{1-1}
phi-3-mini-4k-instruct  &  9  &  15  &  0.42  &  -0.02  &  0.18  &  11.45 \\ \cline{1-1}
gemma-2-27b-it  &  8  &  13  &  0.46  &  -0.18  &  0.0  &  11.49 \\ \cline{1-1}
llama-2-70b-chat-hf  &  7  &  12  &  0.57  &  -0.14  &  0.0  &  11.79 \\ \cline{1-1}
llama-3.1-405B-Instruct-Turbo  &  23  &  57  &  0.23  &  -0.11  &  0.36  &  12.37 \\ \cline{1-1}
llama-3-sonar-large-32k-chat  &  14  &  38  &  0.42  &  0.05  &  0.0  &  13.42 \\ \cline{1-1}
llama-3-70B-Instruct-Turbo  &  22  &  77  &  0.33  &  0.07  &  0.28  &  14.14 \\ \cline{1-1}
llama-3-70B-Instruct-Lite  &  21  &  76  &  0.36  &  -0.0  &  0.0  &  14.73 \\ \cline{1-1}
llama-3-70b-instruct-groq  &  32  &  102  &  0.21  &  -0.13  &  0.43  &  15.36 \\ \cline{1-1}
snowflake-arctic-instruct  &  14  &  69  &  0.76  &  -0.28  &  0.0  &  16.0 \\ \cline{1-1}
c4ai-command-r-plus  &  28  &  139  &  0.37  &  -0.14  &  0.27  &  16.19 \\ \cline{1-1}
mistral-large  &  34  &  153  &  0.27  &  -0.12  &  0.4  &  18.49 \\ \cline{1-1}
qwen2-72B-Instruct  &  34  &  64  &  0.11  &  -0.97  &  0.0  &  22.18 \\ \cline{1-1}
llama-3.1-8B-Instruct-Turbo  &  39  &  77  &  0.1  &  -0.93  &  0.0  &  24.35 \\ \cline{1-1}

\end{tabular}
}
\vspace{0.2cm}
\caption{The topological statistics of the output graphs returned by 21 LLMs when prompted for the Zachary's karate club graph.}
\label{table:stats}
\end{table}

For each examined graph, we denote the set of nodes and of edges with $V$ and $E$ respectively.
We name output graphs in relation with their generating LLM in the first column (in Table \ref{table:stats} the first row being the KC ground truth graph), and provide six relevant statistics for assessing their quality\footnote{We note that the \textit{graph edit distance} \cite{gao2010survey} is not included in these metrics as it becomes intractable for sizes of around 34 nodes, as it can be seen in the output graphs of our sample. This metric is leveraged in Section \ref{s:rank} as we deal with smaller graphs. We present an alternative and more scalable distance metric in Appendix \ref{appendix:spectral}.}. From left to right, we list the number of nodes $|V|$ in the output graph, its number of edges $|E|$, 
its density, assortativity (i.e. tendency of nodes to attach to nodes with similar degrees), and modularity (that is computed w.r.t. the partition provided by a default label propagation on the output graph \cite{networkx}). We argue that modularity is an important indicator considering the frequent use of KC as a benchmark for community detection methods.

Finally we report in the last column the $L2$ distance between the degree distribution of the output graph and KC. Please note that except for rare \textit{unigraphic} graphs, there exists multiple graphs having the same degree distribution, so this metric is not a correctness assessment \emph{per se}. In other words, while a positive distance implies an incorrect result, a distance of 0 does not imply a perfect result (i.e. isomorphic to KC). The table is sorted using the last column.

Table \ref{table:stats} shows that no output graph is exactly correct, i.e. all LLMs hallucinate. The closest being the dbrx-instruct model, with just two hallucinated (i.e. added) edges. ChatGPT models (3.5 and 4o) follow and both return the same output graph. Many outputs are radically different from the correct one, with a surprisingly high heterogeneity among the answers (for instance, with $|E|$ ranging from 8 to 153).   

\begin{figure}[t!]
\centering
\subfloat[qwen2-72B-instr.]{\label{a:misfits-a}\includegraphics[width=0.25\textwidth]{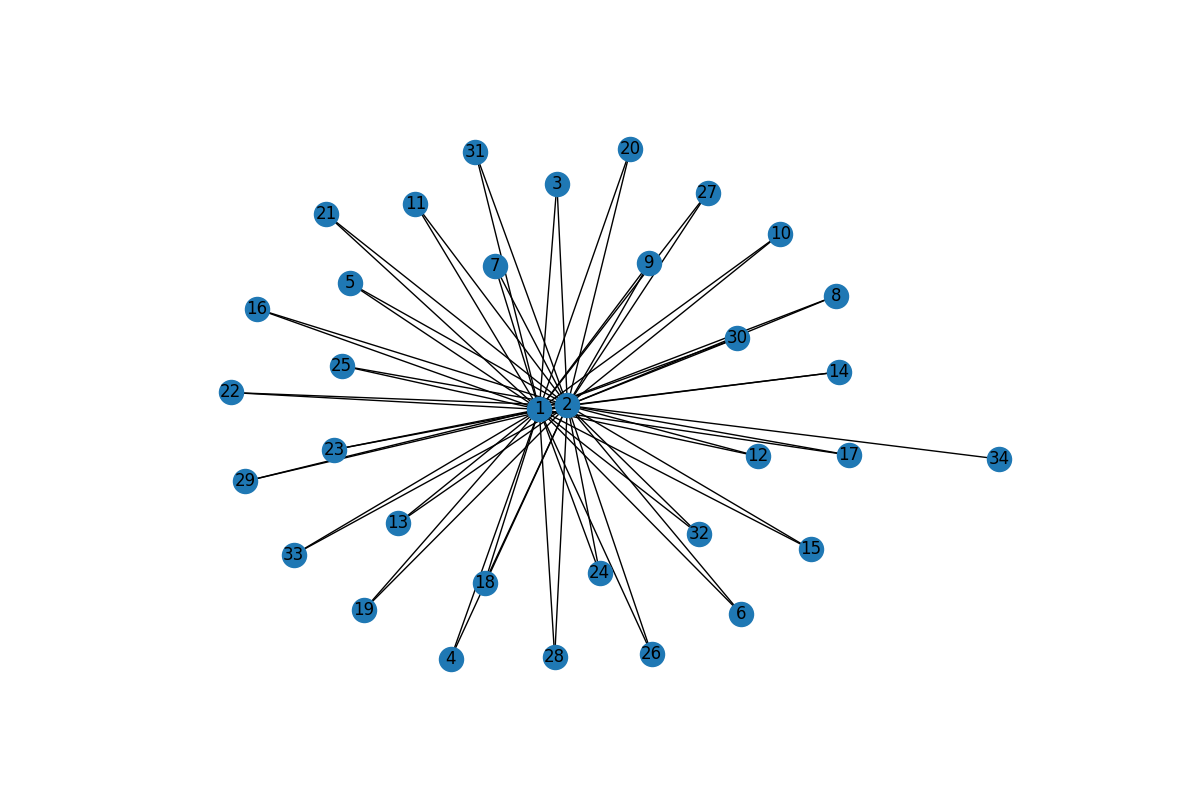}}
\hfill
\subfloat[\scriptsize{llama-2-13b-chat-hf}]{\label{a:misfits-b}\includegraphics[width=0.25\textwidth]{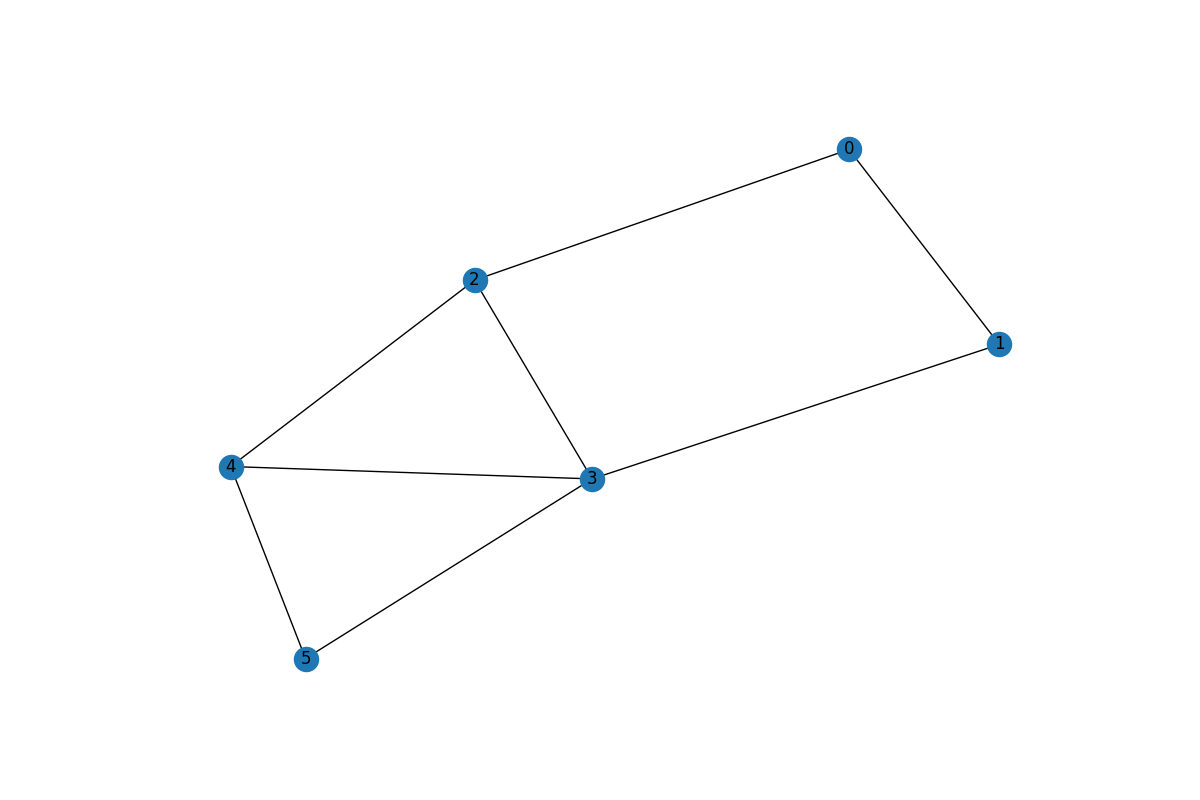}}%
\hfill
\subfloat[mistral-large]{\label{a:misfits-c}\includegraphics[width=0.25\textwidth]{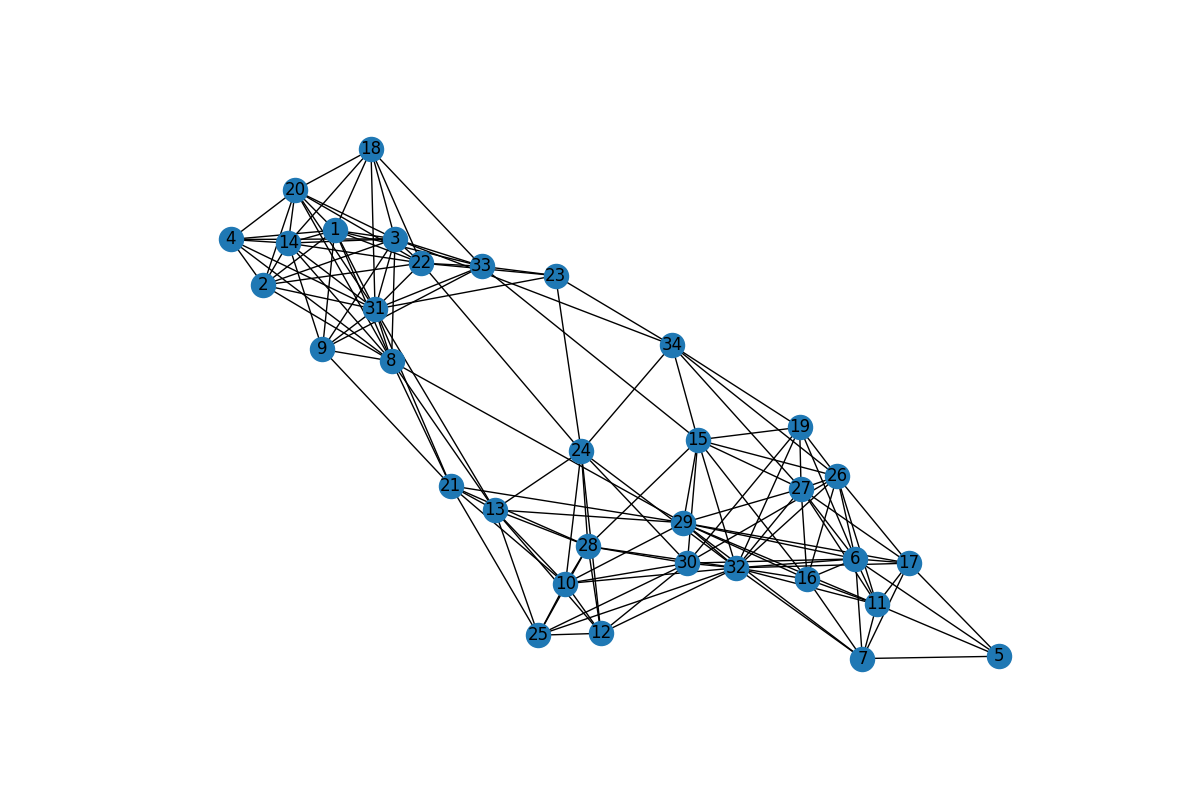}}%
\hfill
\subfloat[\tiny{snowflake-arctic-instr.}]{\label{a:misfits-d}\includegraphics[width=0.25\textwidth]{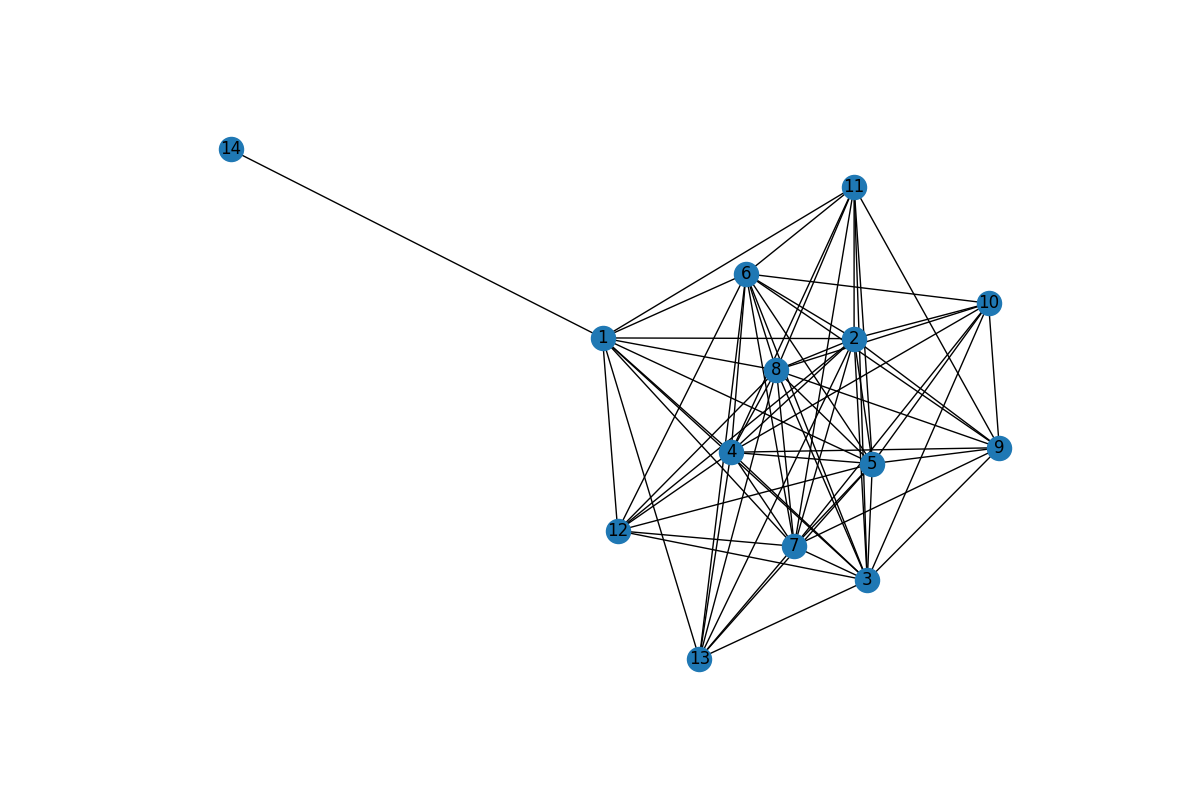}}%
\caption{Example output graphs with salient particularities, when prompted for the Zachary's karate club graph.}
\label{misfits}
\end{figure}

The bottom of the list is populated by output graphs that heavily differ from the ground truth, e.g. the graph on Figure \ref{a:misfits-a} that is star shaped with two central nodes (resulting in a low assortativity score). In this precise case, the number of nodes is correct, 34, and the number of edges close to correct, showing that model has probably learned these global features but not the topology. Figure \ref{a:misfits-b} demonstrates the opposite case for llama-2-13b-chat-hf, where node and edge counts are very low (6 and 8 respectively).
Another interesting case is mistral-large, which generates the correct number of nodes while hallucinating nearly twice the number of edges (see Figure \ref{a:misfits-c}), which explains its large distance to the degree sequence of KC, and its poor modularity in comparison. Finally, Figure \ref{a:misfits-d} showcases for snowflake-arctic-instruct the output of a dense graph (except for a node, resulting in a high density).

When considering families of models, we observe that in the case of the various versions of Llama, a higher number of model parameters does not imply increased correctness. Indeed, the largest Llama, lama-3.1-405B-Instruct-Turbo, has a middle rank. We also observe that for a model version (llama-3-70B), different versions (lite, turbo and groq) result in related ranks, yet with significantly different characteristic outputs.

\subsection{Embedding LLMs Output Graph and their Ground Truth}

We now rely on graph embeddings to evaluate the relative proximity of LLM outputs.
We employ the Karate Club library \cite{karateclub} with the NetLSD method \cite{tsitsulin2018netlsd} with default parameters and Figure \ref{tsne-kc} shows their t-SNE (Les Misérables and graph atlas 50 representations are deferred in Appendix \ref{a:others}.)

\begin{figure}[t!]
\centering
\includegraphics[width=0.7\textwidth]{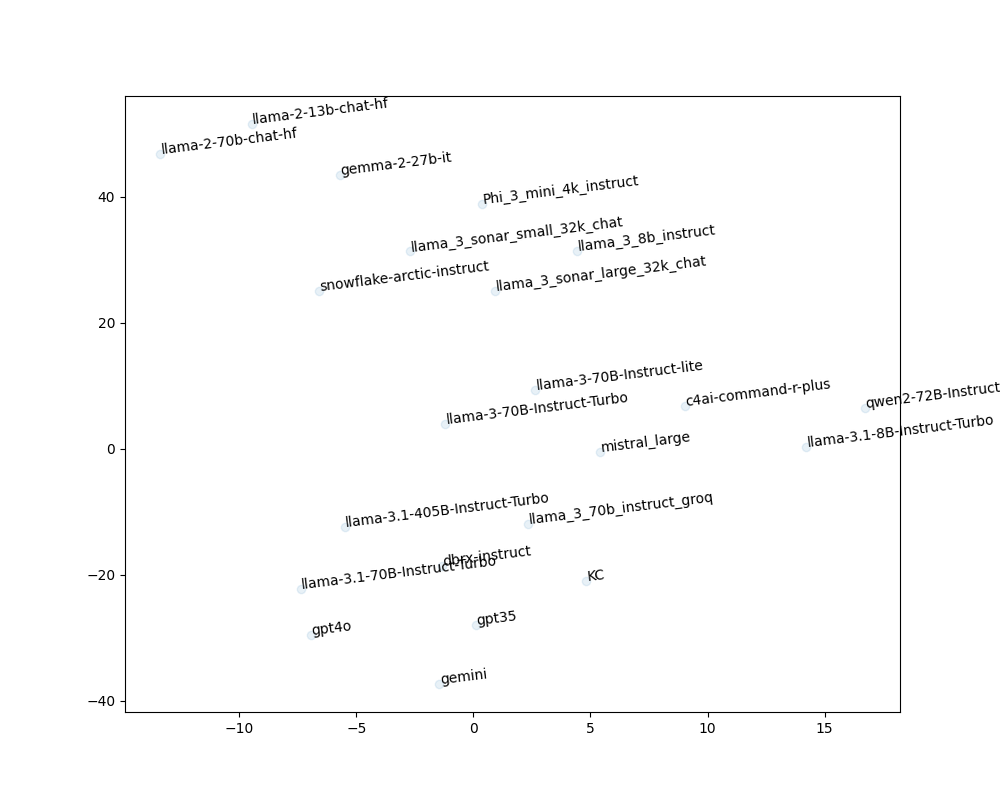}
\vspace{-0.5cm}
\caption{A t-SNE representation of the KC graph and of LLM outputs.}
\label{tsne-kc}
\end{figure}

Despite the proximity of model families such as Llama-2 or ChatGPT, a relatively smooth spread appears in the t-SNE, with clusters that mix graphs belonging to different vendors. Older versions of Llama models (version 2) appear the less related to KC in this embedding.

\section{A LLM Hallucination Ranking Based on Output Graphs}
\label{s:rank}

In this section, we continue to refer to RQ1 and we propose a dedicated metric which allows to rank LLMs based on their hallucination amplitude.
On the other hand, benchmarks for measuring hallucinations are constituted by tens of thousand of prompts \cite{board,medmcqa} and they hence require a privileged access to the model, compared to web browser based prompting access.
Here we propose to sequentially prompt a LLM for just a handful of graphs from the graph atlas, and to average errors made with respect to the reference under a single value (that is the average of graph edit distances).

The graph atlas is composed by $1,252$ different graphs. In our experiment, we choose a resolution of $5$ graphs to be prompted, in particular the first $5$ connected graphs (namely graphs \#3, \#6, \#7, \#13 and \#15). We then compute the exact edit distance of each of these ground truth graphs against their respective LLM output. We finally average these $5$ distances to obtain the final distance score for the queried LLM. We refer to this distance as the Graph Atlas Distance (GAD).

We note that across all tested LLMs and for all datasets, an isomorphism solely occurred for gpt4o on graph atlas \#7 (a triangle graph) and \#13 (a star shaped graph composed of 4 nodes).
The weight given to edit operations are as follows: we do not account for labels, but consider node/edge insertion/deletion, each costing 1.

We compare against the Hallucination Leaderboard \cite{board}, a GitHub page ranking LLMs based on the amplitude of their hallucinations, using a dataset of 50k prompts.
The rank of the 10 tested graphs in common with the Hallucination Leaderboard is presented in Table \ref{table:rank}.

\begin{table}[b!]
\centering
\scriptsize{
\begin{tabular}{|c|c|c|}
    \hline
        \cellcolor[HTML]{DDDDDD}\textbf{Rank} & 
        \cellcolor[HTML]{DDDDDD}\textbf{Graph Atlas Distance (GAD)} & 
        \cellcolor[HTML]{DDDDDD}\textbf{Hallucination Leaderboard$\star$} \\ 
    \hline
    \#1      &   gpt4o (2.2,2.16)                       &  gpt4o    \\
    \#2      &   llama-3.1-70B-Instruct  (6.8,4.76)      &  snowflake-arctic-instruct    \\
    \#3      &   llama-3-8b-chat-hf (8.0,2.34)           &  yi-1.5-34B-Chat   \\
    \#4      &   llama-3.1-405B-Instruct-FP8 (8.2,2.16)  &  llama-3.1-405B-Instruct-FP8  \\
    \#5      &   qwen2-72B-Instruct (8.4,2.3)            &  qwen2-72B-Inst    \\
    \#6      &   yi-1.5-34B-Chat (9.4,2.19)              &  llama-3.1-70B-Instruct   \\
    \#7      &   dbrx-instruct (10.0,5.78)                 &  llama-3-8b-chat-hf      \\
    \#8      &   llama-2-7b-chat-hf (17.2,7.52)          &  c4ai-command-r-plus    \\
    \#9      &   snowflake-arctic-instruct (31.0,24.12)    &  dbrx-instruct    \\
    \#10     &   c4ai-command-r-plus (38.0,41.74)          &  llama-2-7b-chat-hf \\

    \hline
    \multicolumn{3}{|c|}{\textbf{Spearman Correlation: 0.3}} \\
    \hline
    \end{tabular}
}
\vspace{0.2cm}
\caption{GAD (resolution: 5 graphs) vs a summarizing of a benchmark based on prompts with binary answers \cite{board} (ranking from 2024-08-20).
In the GAD column, the first value shows the average edit distance distance, the second the standard deviation.
($\star$ rank in \cite{board} as been adjusted to the LLMs available in our experiment).}
\label{table:rank}
\end{table}

We can observe an interesting yet rather weak correlation in these two rankings (with a Spearman rank correlation of $0.3$, where $0$ is random). The first position is held by gpt4o. The first of the four Llama models (llama-3.1-70B-Instruct) is down-ranked in GAD, yet the three others are in the same order. 
The larger 405B parameters Llama perform worse than smaller models (to be noted that this inversion also appears in \cite{board}, where Llama 3 beats Llama 3.1 with the same amount of parameters (70B)). qwen2-72B-Instruct similarly ranks in the middle, while c4ai-command-r-plus ranks at the bottom. snowflake-arctic-instruct is nevertheless strongly down-ranked with GAD, as compared to its second position in the Hallucination Leaderboard.

In the light of the fact that we are using only five graphs to evaluate an LLM, when comparing to the use of 50k prompts in the Hallucination Leaderboard, we find this result to be encouraging for further study on the discriminative power of querying LLMs for structured data such as graphs.

\section{Ability to Generate Random Graphs (RQ2)}

We now investigate the ability of LLMs to generate randomness in a structured form. We choose to focus on the generation of \er random graph. 
As a prerequisite for generating random graphs, a first natural question is whether LLMs can generate any randomness. Recent literature provides a positive answer: 
\citet{hopkins2023can} looked at random number generation 
and shows that it is possible in some settings and for some LLMs to generate a rather uniform distribution on a small interval. On a different perspective, \citet{vankoevering2024randomrandomevaluatingrandomness} show that LLMs are keen to replicate human biases in random generations and at times even exacerbate them. On the other hand, Harrison \citet{harrison2024comparisonlargelanguagemodel} finds that gpt3.5 has lower repeat frequencies and adjacent number frequencies than humans.

In this work, our goal is to evaluate the randomness ability through the concept of the emergent ability and the lens of hallucination (which will connect to RQ1). To support this rationale, a simple baseline is desirable to evaluate this generative task. Consequently, we choose to focus on \er random graphs, as it is arguably the simplest generative model from the literature (see e.g. Watts-Strogatz \cite{watts1998collective} or Barabasi-Albert \cite{albert2002statistical} models for an increased amount of parameters). 

\subsection{Experimental Method}

\paragraph{Selecting LLMs and prompts.}
In RQ1, we issue a single request to LLMs to obtain the recited graph. In RQ2, as the outputs are randomized, we now require more requests to judge the quality of the generated graphs. For this reason, we choose to investigate to open weight LLMs that we run on the Jean Zay supercomputer\footnote{http://www.idris.fr/jean-zay/}. We list the 25 selected LLMs in Figure \ref{figs: F3}.

\paragraph{Prompting strategy}
We explore two prompting strategies. The first strategy asks straightforwardly for a \er graph realization and specifies the format of the sought output. We refer to this prompt as \textit{Direct Prompt}. As prompts are known to greatly influence the LLM outputs, we also experiment with a Chain of Thought prompt \textit{CoT} that generally elicits better results \cite{wei2023chainofthoughtpromptingelicitsreasoning}. The precise prompts are described in appendix \ref{appendix: Experiment settings}.

\subsection{Evaluating LLM Outputs}
We now need to assess the quality of the generated outputs. We are interested in knowing if the generated graphs are representative of \er graphs. To this end, we resort to a graph-level hypothesis testing strategy.

\paragraph{Parsing the generated outputs.}
As in RQ1 setting, we leverage a regular expression based script to extract the edge list from the LLM output (see appendix \ref{appendix: Experiment settings} for more details). 
We define $\mathcal{G}(M, T, n, p)$ as the set of graphs that have been \emph{successfully} extracted from the set of queries $N_{\text{queries}}$ given to the LLM $M$, with temperature $T$\footnote{The temperature of an LLM is a sampling parameter that can be set at the inference. Increasing $T$ flattens the next token probability distribution over the token vocabulary, while having $T\simeq 0$ forces the model to having a deterministic output \cite{zhu2024hot}.} and $(n,p)$ being the \er parameters (with $n$ the number of nodes, and $p$ the probability that an edge exists between any two distinct nodes).
We precise that we refer to the graphs that are ``successfully" extracted, because, as in the experiments from RQ1, LLMs might not always respect the demanded format in their outputs. If this happens, our regular expression may fail to identify a graph within the LLM output. Let 
\begin{equation}
    \sigma(M,T,n,p) \DEF \frac{|\mathcal{G}(M, T, n, p)|}{N_{\text{queries}}}
\end{equation}
be the fraction of \textit{Syntactically Correct Answers}, at least with respect to our regular expression based algorithm. In our experiments, $N_{\text{queries}}$ is set to $200$.

\paragraph{Hypothesis Testing.}

Let $H_0$ be: ``The graph has been generated from an \er process". As this (graph-level) hypothesis is not easy to test, we resort to a second (node-level) hypothesis $\tilde{H_0}$ whose test is more accessible. To construct $\tilde{H_0}$ we rely on a property of the \er model: as a consequence of the graph generation process in which each edge is added independently, node degrees follow a binomial distribution. We hence define $\tilde{H_0}$: ``Node degrees follow a binomial distribution", the rationale being that since $H_0 \Rightarrow \tilde{H_0}$, rejecting $\tilde{H_0}$ leads to rejecting $H_0$.

\paragraph{Metric definition.}

We use a $\chi^2$ hypothesis test, whose statistic is:

\begin{equation}\label{eq: Pearson's N_graphs}
\chi^2=\sum_{i=0}^{n-1} \frac{(d_i - n\beta_i)^2}{n\beta_i},
\end{equation}    
where $(d_i)_{0\leq i \leq n-1}$ is the number of graph nodes having $i$ neighbors 
and $\beta_i=\mathbb{P}(X=i)$ (with $X\sim Bin(n-1,p)$ binomial distribution and $X$ a random variable).

We transform this test statistic into an hypothesis testing by comparing the obtained value $\chi^2$ to the canonical Chi-square distribution 
having $n$ degrees of freedom. This yields a \text{p-value} with which we now define, for a graph $g$, a success: $\chi^2_\text{test}(g) \DEF \ONE_{\text{p-value} > \alpha}$.
In other words, given a graph $g$ we consider that $g$ succeeds our $\chi^2$ test if the hypothesis that its degree distribution follows a binomial distribution cannot be rejected with confidence $\alpha$.
We will use $\alpha=0.001$ throughout the paper; this is a pretty low value, which makes this test 
a lenient metric that only detects highly inaccurate graphs that arguably deserve being called hallucinations.

Note that this test is not sufficient, since it allows to conclude that graphs failing this test are not generated from an \er model with a good confidence. On the other hand, it does not guarantee that graphs passing this test come indeed from an \er model. 
In fact, it could be possible to have a graph generation processes that would result in graphs having binomial degree distribution but that are not \er.

We can define the Test Success Rate for a set of graphs $\mathcal{G}$ as the average $\chi^2$ test success rate:
\begin{equation}
  \gamma(M,T,n,p) \DEF 1/|\mathcal{G|}{}\sum_{g \in \mathcal{G}} \chi^2_\text{test}(g) \in [0,1]
\end{equation}

Note that this metric needs to have a fixed number of nodes $n$ and knowledge of the distribution of $X$ (to determine $\beta_i$), in our case the binomial distribution for \er graphs.

As a baseline for this metric, we computed the performance on graphs generated by the algorithm for the \er model from the Python library NetworkX). This will be labeled as the ``True Erdős–Rényi model". We note that the $\gamma$ metric is not perfect, as NetworkX achieves an average score of 97\% (instead of an ideal 100\%). This gap only appears for small values of $n$, which is predictable since $n$ directly determines the number of samples for each $\chi^2_{\text{test}}$, since there are as much samples as many node degrees. The low number of samples weakens the test, thus explaining the observed gap.

\begin{figure}[t!]
\centering
\includegraphics[width=0.8\textwidth]{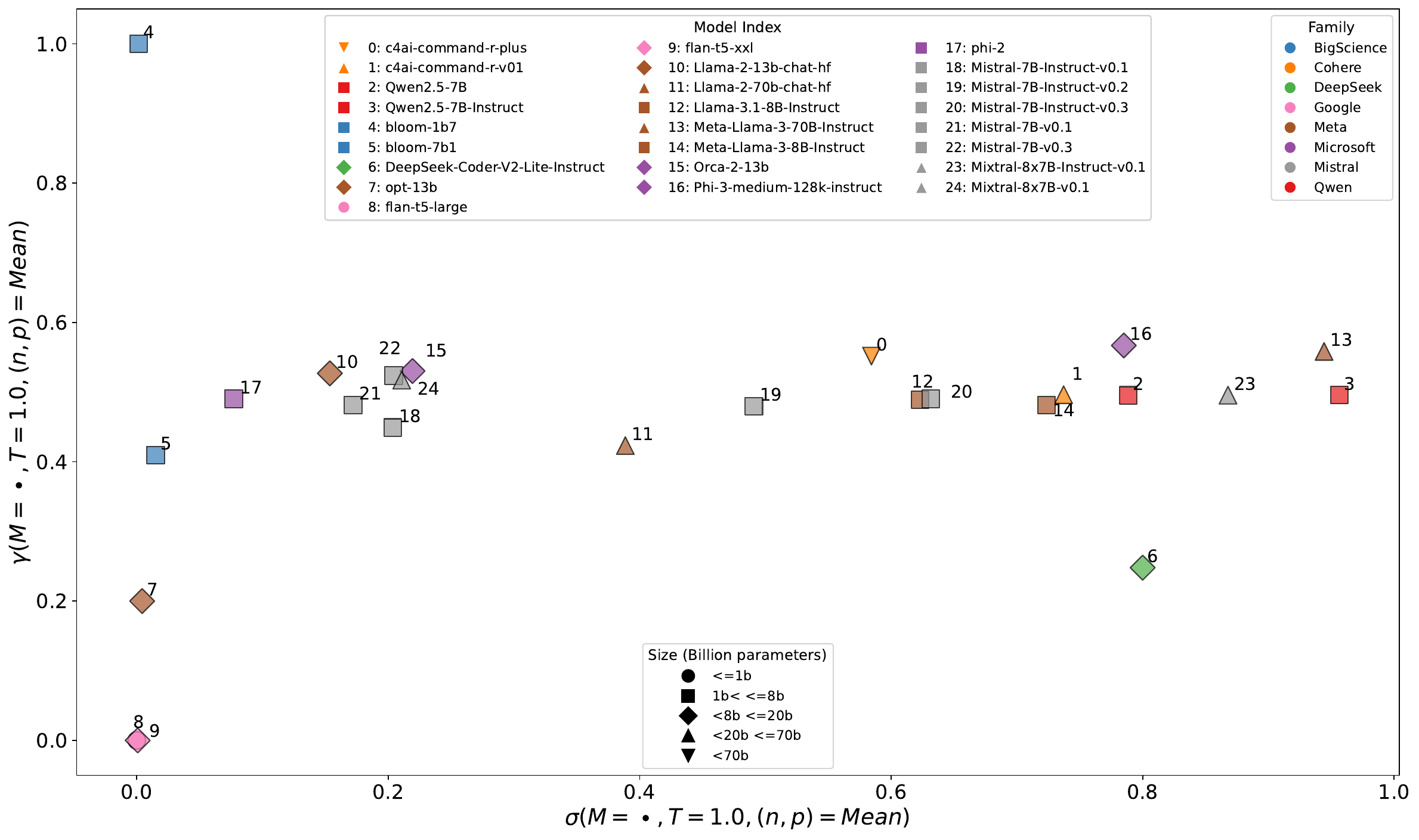}
\vspace{-0.1cm}
\caption{Test Success Rate $\gamma$ and Syntactically Correct Answer Rate $\sigma$ over LLM's outputs on \er graphs. The 7 $(n,p)$ Erdős–Rényi parameters are aggregated by mean. The experiment includes 25 LLMs and 200 graph queries per LLM-$(n,p)$ pair. Temperature is set to 1.0 and we are here using the CoT Prompt.}
\label{figs: F1_all_param}
\end{figure}

\subsection{Experiments on the Generative Capabilities of 25 LLMs}

Figure \ref{figs: F1_all_param} represents simultaneously the syntactically correct answer rate $\sigma$ and the performance $\gamma$, given that both constitute important criteria when assessing the ability of LLMs to generate random graphs. 
The majority of models has a performance $\gamma$ around 45\%.
However the syntactically correct answers rate $\sigma$ is much more heterogeneous, with two major model clusters: one under 25\% and the other above 60\%. Finally five models (\#4, \#5, \#7, \#8, \#9) have a really low $\sigma$, and two of them (\#8, \#9) fail completely the test.

We observe that there is no generalized impact of LLMs belonging to a given family of models, and no impact of the model size. 
However there are some model pairs of family and size that are close (\#19, \#20), (\#2, \#3), (\#12, \#14); though some others model pairs of family and size are far away (\#21, \#20), (\#23, \#24). We finally note that the \textit{Instruct} models generally work better that non \textit{Instruct} ones, as it was expected given that they have been finetuned to better respond to natural language instructions.

\begin{figure}[t!]
\centering
\includegraphics[width=0.8\textwidth]{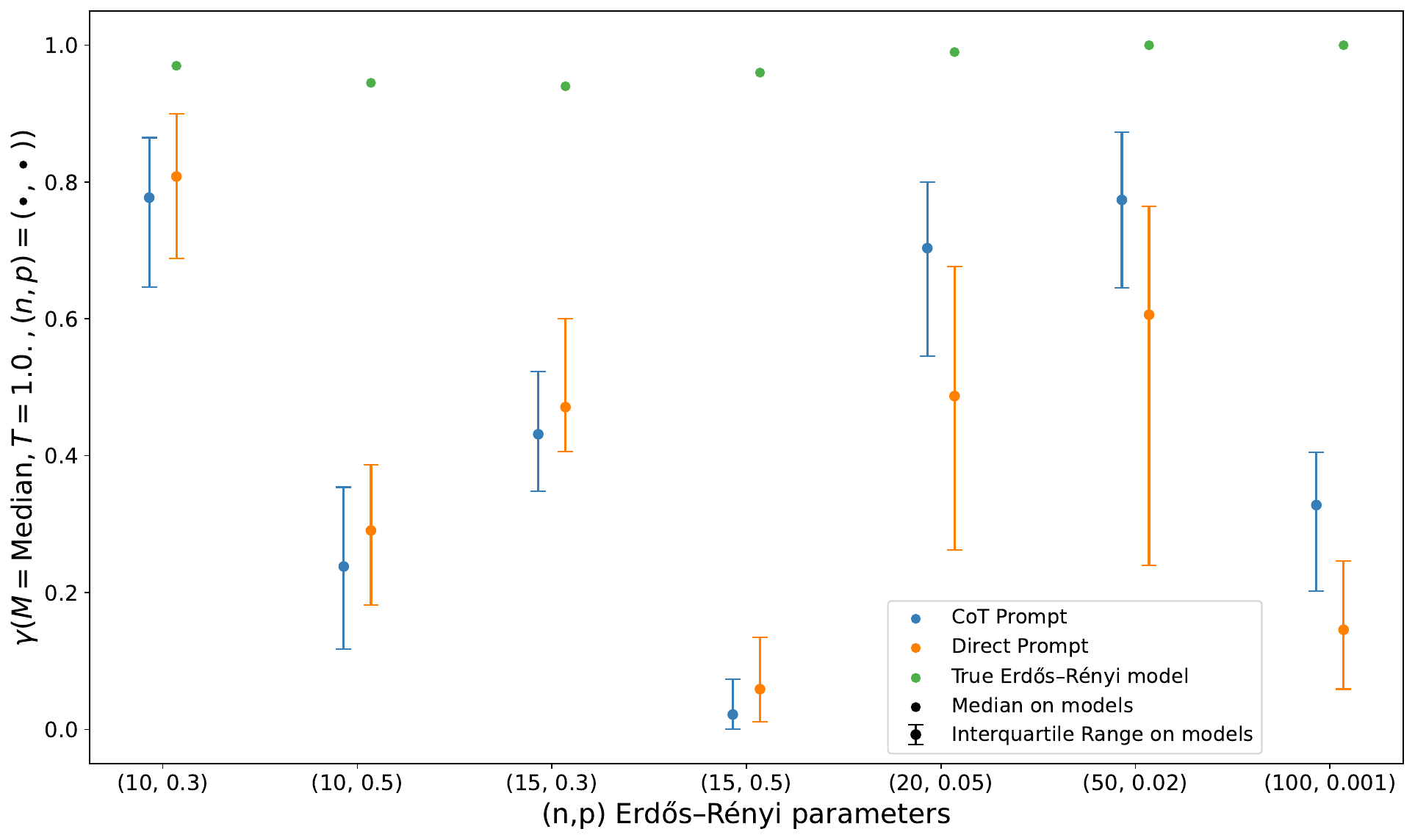}
\vspace{-0.1cm}
\caption{Erdős–Rényi parameters $(n,p)$ influence on Test Success Rate $\gamma$, with prompt variation. The models $M$ are represented as median and interquartiles. The experiment includes 25 LLMs, 7 $(n,p)$ parameter pairs and 200 graph queries per LLM-$(n,p)$-prompt triplet. Temperature is 1.0.}
\label{figs: F3}
\end{figure}

Figure \ref{figs: F1_all_param} averages out the impact of different Erdős–Rényi parameters, hiding the variability they cause.
The variability is highlighted in Figure~\ref{figs: F3}, which shows the median performance of models on different parameters and the interquartile range, in order to underline the amplitude of the variability across models. The apparent uniformity in Figure~\ref{figs: F1_all_param} should be interpreted with caution, as there are significant variations across both parameter pairs and LLMs.
Interestingly, Figure~\ref{figs: F3} shows that all models struggle to correctly produce certain \er parameter pairs, with $(n=15, p=0.5)$ being the hardest. 

We see that for the parameter pairs $(n, p\leq 1/n)$ (on the rightmost side), the \textit{CoT Prompt} performs significantly better than the \textit{Direct Prompt}. In contrast, for the other parameter pairs, the performance is more balanced, with the \textit{Direct Prompt} performing slightly better.

\begin{figure}[t!]
\centering
\includegraphics[width=0.8\textwidth]{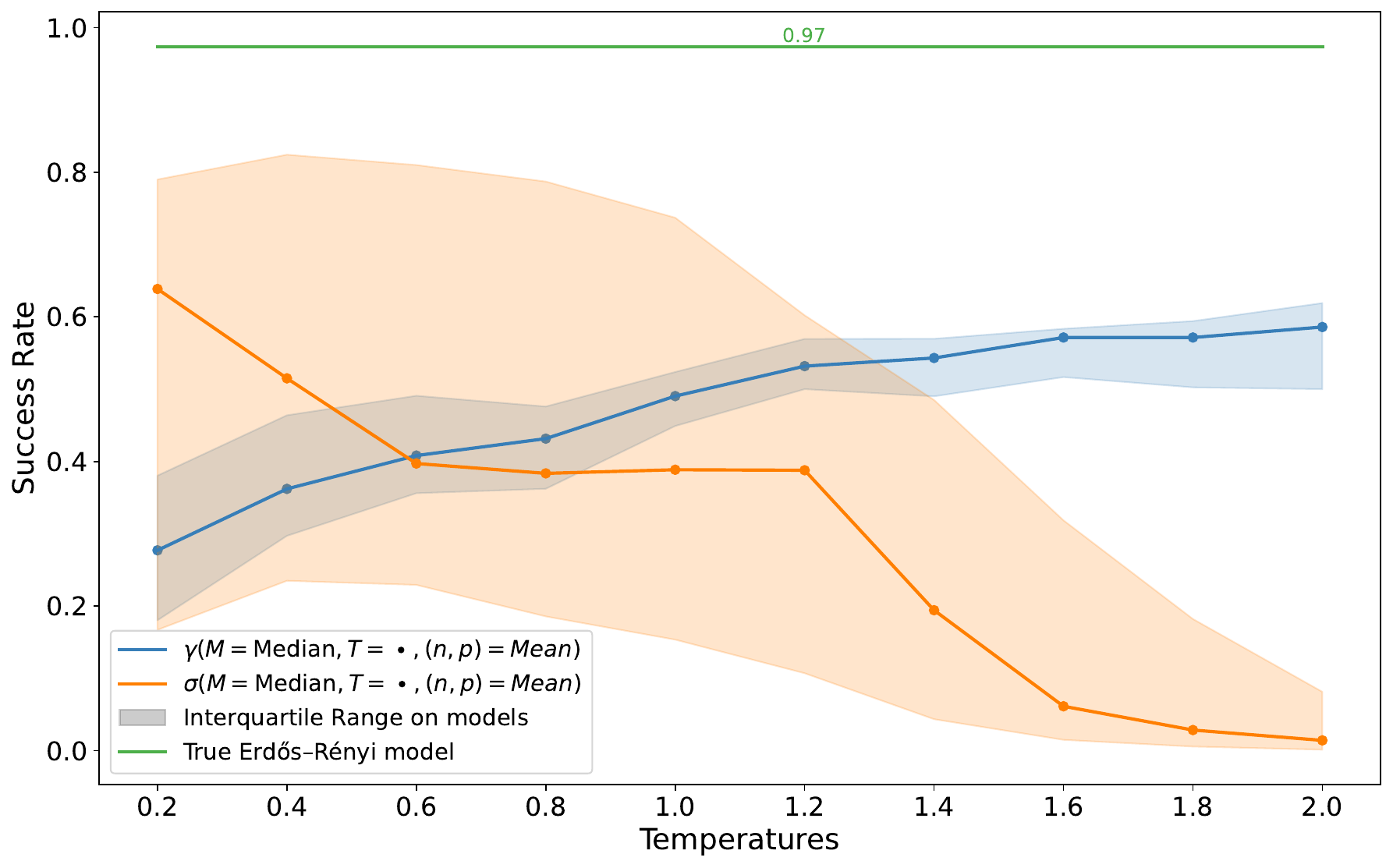}
\vspace{-0.1cm}
\caption{Temperature influence on (degree distribution based) Test Success Rate $\gamma$ and Syntactically Correct Answer Rate $\sigma$ on \er graphs queried. The $(n,p)$ Erdős–Rényi parameters dimension is averaged while the models $M$ are represented as median and interquartiles. The experiment includes 25 LLMs, 7 $(n,p)$ parameter pairs and 200 graph queries per LLM-Temperature-$(n,p)$ triplet.}
\label{figs: F2}
\end{figure}

Furthermore, Figure \ref{figs: F2} represents the impact of the LLM temperature $T$ on the task. As an increase in temperature flattens the probability distribution of the next token in the output of a LLM \cite{zhu2024hot}, it raises the likelihood of ending up generating incoherent text. Consequently, the decreasing trend in the curve of syntactically correct answers (orange) is expected.
In contrast, increasing the temperature interestingly improves the success rate $\gamma$.
It should be noted that this curve is averaged over the seven tested parameter pairs $(n,p)$, and that there exist some parameter pairs for which the behavior differs. 

Finally, Table~\ref{table:Hallu_random_rank} shows correlation with the Hallucination Leaderboard \cite{board2}, yet a weaker correlation than what found in RQ1 (see Table \ref{table:rank}), with a Spearman correlation of $0.2$. On the other hand, we can see that in both cases Orca-2-13b is ranked first, Qwen is ranked high, and phi-2 sits in the middle of both rankings.

\begin{table}[t!]
\centering
\scriptsize{
\begin{tabular}{|c|c|c|}
    \hline
        \cellcolor[HTML]{DDDDDD}\textbf{Rank} & 
        \cellcolor[HTML]{DDDDDD}\textbf{Test Success Rate $\gamma$ (ours)} & 
        \cellcolor[HTML]{DDDDDD}\textbf{Hallucination Leaderboard$\star$} \\ 
    \hline
    \#1 & microsoft/Orca-2-13b & microsoft/Orca-2-13b \\ 
\#2 & meta-llama/Llama-2-13b-chat-hf & Qwen/Qwen2.5-7B-Instruct \\ 
\#3 & Qwen/Qwen2.5-7B-Instruct & meta-llama/Llama-2-70b-chat-hf \\ 
\#4 & mistralai/Mixtral-8x7B-Instruct-v0.1 & microsoft/phi-2 \\ 
\#5 & microsoft/phi-2 & mistralai/Mistral-7B-Instruct-v0.3 \\ 
\#6 & mistralai/Mistral-7B-Instruct-v0.3 & meta-llama/Llama-2-13b-chat-hf \\ 
\#7 & meta-llama/Llama-2-70b-chat-hf & mistralai/Mixtral-8x7B-Instruct-v0.1 \\ 

    \hline
    \multicolumn{3}{|c|}{\textbf{Spearman Correlation: 0.2}} \\
    \hline
    \end{tabular}
}
\vspace{0.2cm}
\caption{
Our Test Success Rate $\gamma$ vs a summarizing of a benchmark based on prompts with binary answers \cite{board2} (ranking from 2025-03-20). 
To be noted that $\star$ rank in \cite{board2} as been adjusted to the LLMs available in our experiment.
}
\label{table:Hallu_random_rank}
\end{table}
\paragraph{Discussion.}
Despite the lenient nature of our metric for evaluating the randomness quality, the magnitude of its success suggests that an emergent ability is likely present. However, this interpretation must be considered with the remaining uncertainty on the number of samples in their training data, which could imply that the observed performance is merely a result of memorization.

Also, note that with LLM experiments, there are many parameters that can have significant impact on the results. They are essentially sampling parameters (temperature, top-k etc.), and the prompting strategy \cite{mizrahi2024stateartmultipromptllm} \cite{do2024automaticpromptselectionlarge}. Hence, our results offer a lower bound on the LLMs' ability on the chosen task. Note that the models in the bottom left of Figure~\ref{figs: F1_all_param}, which perform poorly, might be more successful in different tasks other than the one tested in RQ2 \cite{mizrahi2024stateartmultipromptllm}.

\paragraph{Gathering raw information, a meaningful abstraction?}
Both RQ1 and RQ2 explore queries from graphs as a way to obtain more information bits per prompt, compared to the standard MCQ approach in which each request yields a number of bits capturing the number of possible answers to each question (typically 2 or 4 choices, hence 1 or 2 bits). 
Implicitly, RQ1 compares interrogation patterns ``bit to bit", regardless of the relevance of the collected bits. A natural limitation is to overlook the precise meaning of each bit. Concretely, when one bit yielded by our approach captures whether the target LLM correctly predicted a (Jean Valjean, Cosette) relation in Les Misérables, one bit yielded by MedMCQA benchmarks captures whether the target LLM correctly associates ``A 40-year-old man has megaloblastic anemia and early signs of neurological abnormality" to a deficit in B12 Vitamin. In RQ2, while a bit produced in a random graph is not very useful on its own right, when enough are collected, we are able to form a group that can be statistically evaluated.
The difficulty with considering the semantics associated with each bit is that the notion of relevance is strongly application dependent. The relevance varies depending on the use case, 
for instance when evaluating the ability to provide truthful medical advice.

\section{Related Work}
Graphs are already used in various ways when it comes to studying the LLMs hallucinations \cite{nadeau2024benchmarkingllama2mistralgemma,tonmoy2024comprehensivesurveyhallucinationmitigation,sansford2024grapheval}, in order to  evaluate the quality of their outputs.

\textit{Knowledge graphs} are leveraged in a LLM-based hallucination evaluation framework \cite{sansford2024grapheval}, by prompting for text and checking the correctness of the output having a binary labeled dataset at hand. Knowledge graphs are constructed from unstructured textual data by identifying the set of entities within the text and the relationships between them, resulting in a structured representation.
Work in \cite{hao2024quantifying} models social networks as graphs on which they simulate information spreading in order to track LLM hallucinations flowing within these networks.
Nonkes et al. \cite{nonkes2024leveraging} create a graph structure that connects generations that lie closely in the embedding space of hallucinated and non-hallucinated LLM text generations. Graph Attention Networks then learn this structure and generalize it to unseen generations for categorization.
Work in the complex networks community \cite{jeyaram2023reconstructing} is interested in extracting organizational networks (bipartite user-activity networks here) from raw text obtained from LLMs, e.g. in the context of standardized individual-level contributions across a large numbers of teams participating to a competition. 

\citet{hopkins2023can} investigated random number generation capabilities of LLM. They identified several factors that impact the generation quality, and show that in some conditions, some LLMs may generate rather uniform distributions on small intervals. \citet{vankoevering2024randomrandomevaluatingrandomness} show that LLMs are keen to replicate  and even exacerbate human biases in random generations.

In complex network analysis, and to evaluate the quality of \er graphs, \citet{ouadah2019degreebasedgoodnessoffittestsheterogeneous} propose an approach based on degree variance goodness-of-fit, but their model of \emph{heterogeneous} \er law does not apply to our setting.

Finally, the concept of emergent ability has two definitions in literature. We chose to focus on one which is the \textit{ability to reason about novel problems zero-shot, without any direct training} (\citet{webb2023emergent}). The other being an \textit{ability that is
not present in smaller-scale models but is present in large-scale models} (\citet{wei2022emergent}), that focus on the scaling complexity of the neural networks eliciting ability while the first definition focus on the ability to perform a certain task for which no example was present in the training data.
Also, \citet{schaeffer2023emergent} challenges the idea that LLMs develop novel abilities suddenly as their scale increases. The authors argue that these abilities appear emergent due to specific measurement and analysis methods rather than fundamental changes in the models.
To be best of our knowledge, our work is the first to propose a head-to-head comparison of a ground truth graph to a prompted output graph, or a statistical evaluation of the generative capabilities of LLMs. This was performed under the scope of hallucinations, and allows for novel assessments.

\section{Conclusion}

The scope of this article is to compare the ability of LLMs to (1) recite ground truth graphs from the literature and (2) generate Erdős–Rényi random graphs. A first striking observation is that current LLMs are far from being reliable sources (i.e. they do hallucinate graphs) in both tasks. 
Nevertheless, these glitches open solid comparison avenues, which we exploited to observe the significant differences of LLMs in their hallucination amplitude regarding standard graphs.
We introduced two relevant metrics for the tasks and we showed that a handful of prompts can correlate with a method leveraging tens of thousands of queries for binary answers, with the goal to rank the LLMs that hallucinating the least. For what concerns the ability to generate random graphs, the generative quality we measured is high enough to raise intriguing questions and motivate deeper exploration, e.g. starting with an extension to other generative models.
We believe that such a novel perspective leaves ways for future work to better evaluate LLMs on both hallucination and randomness generative abilities.

\bibliographystyle{plainnat}  
\bibliography{references}

\section{Acknowledgements}

This work was granted access to the HPC resources of IDRIS under the allocation AD011015776 made by GENCI. We thank the French Jean Zay platform, used for the computations required for RQ2.

This research is supported by French ANR PACMAM (ANR-24-CE23-7787).

\appendix

\section{Example Output From a LLM}
\label{appendix:output}

We collected this example at the following url on the together.ai platform:
{\scriptsize \url{https://api.together.xyz/playground/chat/meta-llama/Meta-Llama-3.1-405B-Instruct-Turbo}}, with the following prompt: \texttt{Provide me the so called "Zachary's karate club" graph as a python edge list; print it}.
Here is the full response from that model:

\begin{quote}
Zachary's Karate Club is a well-known social network dataset that represents the friendships between 34 members of a karate club at a US university in the 1970s. The graph is often used as a benchmark for community detection algorithms.

Here is the graph represented as a Python edge list:

\scriptsize{
\begin{verbatim}
import networkx as nx

# Define the edge list
edges = [
    (1, 2), (1, 3), (2, 3), (1, 4), (2, 4), (3, 4), (1, 5), (2, 5), (1, 6), (3, 6),
    (1, 7), (5, 7), (6, 7), (1, 8), (2, 8), (3, 8), (1, 9), (3, 9), (3, 10), (3, 11),
    (1, 12), (1, 13), (4, 13), (3, 14), (3, 15), (6, 16), (6, 17), (7, 17), (6, 18),
    (1, 20), (2, 20), (1, 22), (2, 22), (24, 26), (25, 26), (3, 28), (3, 29), (3, 30),
    (24, 31), (27, 31), (2, 32), (24, 32), (2, 33), (4, 33), (24, 33), (25, 33),
    (26, 33), (27, 33), (28, 33), (29, 33), (30, 33), (31, 33), (32, 33)
]

# Create a graph from the edge list
G = nx.Graph(edges)

# Print the graph
print("Nodes:", G.nodes())
print("Edges:", G.edges())
\end{verbatim}
}
This code defines the edge list for Zachary's Karate Club graph and creates a NetworkX graph object from it. The print statements display the nodes and edges of the graph.

Note that the node labels are integers from 1 to 34, and the edges represent friendships between the
\end{quote}

Note that the response is cut as presented here (default output length on this platform is 512).

\section{Statistics and Embeddings for Les Misérables and Graph Atlas 50}
\label{a:others}

\begin{table}[h!]
\centering
\scriptsize{
\begin{tabular}{|c|cccccc}
\hline
\cellcolor[HTML]{DDDDDD}\textbf{LLM} & \multicolumn{1}{c|}{\cellcolor[HTML]{DDDDDD}\textbf{|N|}} & \multicolumn{1}{c|}{\cellcolor[HTML]{DDDDDD}\textbf{|V|}} & \multicolumn{1}{c|}{\cellcolor[HTML]{DDDDDD}\textbf{density}} & \multicolumn{1}{c|}{\cellcolor[HTML]{DDDDDD}\textbf{assort.}} & \multicolumn{1}{c|}{\cellcolor[HTML]{DDDDDD}\textbf{modularity}}  & \multicolumn{1}{c|}{\cellcolor[HTML]{DDDDDD}\textbf{dist. to LM deg. seq.}} \\ \hline
(reference: Les Misérables)\cellcolor[HTML]{EEEEEE}  &  77 \cellcolor[HTML]{EEEEEE} &  254\cellcolor[HTML]{EEEEEE}  &  0.09\cellcolor[HTML]{EEEEEE}  &  -0.17 \cellcolor[HTML]{EEEEEE} &  0.46\cellcolor[HTML]{EEEEEE}  &  0.0 \cellcolor[HTML]{EEEEEE}\\ \cline{1-1}

gpt4o  &  66  &  180  &  0.08  &  -0.22  &  0.5  &  7.55 \\ \cline{1-1}
c4ai-command-r-plus  &  55  &  87  &  0.06  &  -0.57  &  0.11  &  17.49 \\ \cline{1-1}
llama-3.1-405B-Instruct-Turbo  &  26  &  38  &  0.12  &  -0.56  &  0.63  &  17.64 \\ \cline{1-1}
qwen2-72B-Instruct  &  25  &  33  &  0.11  &  -0.72  &  0.0  &  18.33 \\ \cline{1-1}
dbrx-instruct  &  9  &  11  &  0.31  &  0.16  &  0.29  &  22.76 \\ \cline{1-1}
gemini  &  14  &  15  &  0.16  &  -0.02  &  0.37  &  22.07 \\ \cline{1-1}
mistral-large  &  15  &  19  &  0.18  &  0.08  &  0.43  &  22.09 \\ \cline{1-1}
gemma-2-27b-it  &  14  &  25  &  0.27  &  -0.1  &  0.4  &  23.56 \\ \cline{1-1}
llama-3-70b-instruct-groq  &  24  &  47  &  0.17  &  0.11  &  0.59  &  23.35 \\ \cline{1-1}
phi-3-mini-4k-instruct  &  9  &  12  &  0.33  &  -0.21  &  0.21  &  23.07 \\ \cline{1-1}

\end{tabular}
}
\caption{Statistics for graphs returned by 10 LLMs (Les Misérables).}
\label{table:stats-lm}
\end{table}

\begin{table}[h!]
\centering
\scriptsize{
\begin{tabular}{|c|cccccc}
\hline
\cellcolor[HTML]{DDDDDD}\textbf{LLM} & \multicolumn{1}{c|}{\cellcolor[HTML]{DDDDDD}\textbf{|N|}} & \multicolumn{1}{c|}{\cellcolor[HTML]{DDDDDD}\textbf{|V|}} & \multicolumn{1}{c|}{\cellcolor[HTML]{DDDDDD}\textbf{density}} & \multicolumn{1}{c|}{\cellcolor[HTML]{DDDDDD}\textbf{assort.}} & \multicolumn{1}{c|}{\cellcolor[HTML]{DDDDDD}\textbf{modularity}}  & \multicolumn{1}{c|}{\cellcolor[HTML]{DDDDDD}\textbf{dist. to ga50 deg. seq.}} \\ \hline
(reference: graph atlas 50)\cellcolor[HTML]{EEEEEE} & 5 \cellcolor[HTML]{EEEEEE} &  8\cellcolor[HTML]{EEEEEE}  &  0.8\cellcolor[HTML]{EEEEEE}  &  -0.33 \cellcolor[HTML]{EEEEEE} &  0.0 \cellcolor[HTML]{EEEEEE} &  0.0\cellcolor[HTML]{EEEEEE} \\ \cline{1-1}
gemma-2-27b-it  &  5  &  7  &  0.7  &  -0.5  &  0.0  &  2.83 \\ \cline{1-1}
gpt4o  &  6  &  9  &  0.6  &  1.0  &  0.0  &  2.24 \\ \cline{1-1}
llama-3.1-405B-Instruct-Turbo  &  5  &  6  &  0.6  &  -0.29  &  0.0  &  2.0 \\ \cline{1-1}
c4ai-command-r-plus  &  5  &  10  &  1.0  &  1.0  &  0.0  &  5.66 \\ \cline{1-1}
llama-3-70b-instruct-groq  &  12  &  13  &  0.2  &  0.21  &  0.37  &  6.4 \\ \cline{1-1}
dbrx-instruct  &  10  &  45  &  1.0  &  1.0  &  0.0  &  10.82 \\ \cline{1-1}
qwen2-72B-Instruct  &  58  &  57  &  0.03  &  -0.26  &  0.65  &  37.22 \\ \cline{1-1}
mistral-large  &  100  &  375  &  0.08  &  -0.5  &  0.8  &  70.83 \\ \cline{1-1}
\end{tabular}
}
\caption{Statistics for graphs returned by 8 LLMs (graph atlas \#50).}
\label{table:stats-ga50}
\end{table}

\begin{figure}[h!]
\centering
\subfloat[Les Misérables graph]{\label{4figs-a}\includegraphics[width=0.49\textwidth]{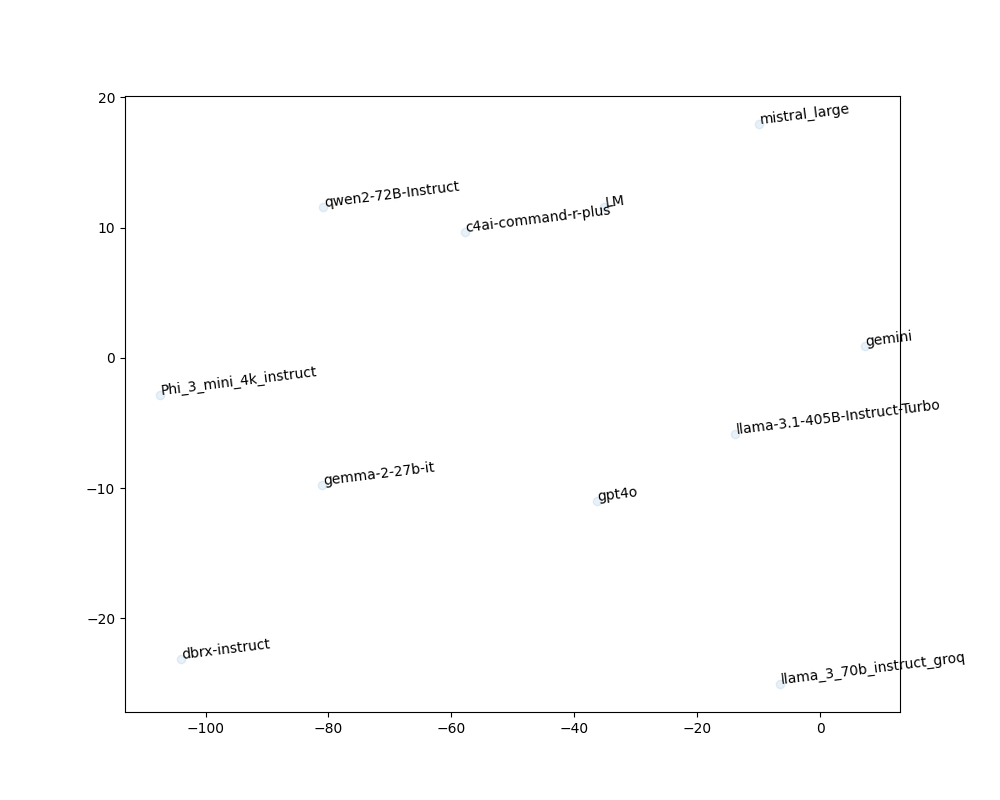}}
\hfill
\subfloat[Graph Atlas 50]{\label{4figs-b}\includegraphics[width=0.49\textwidth]{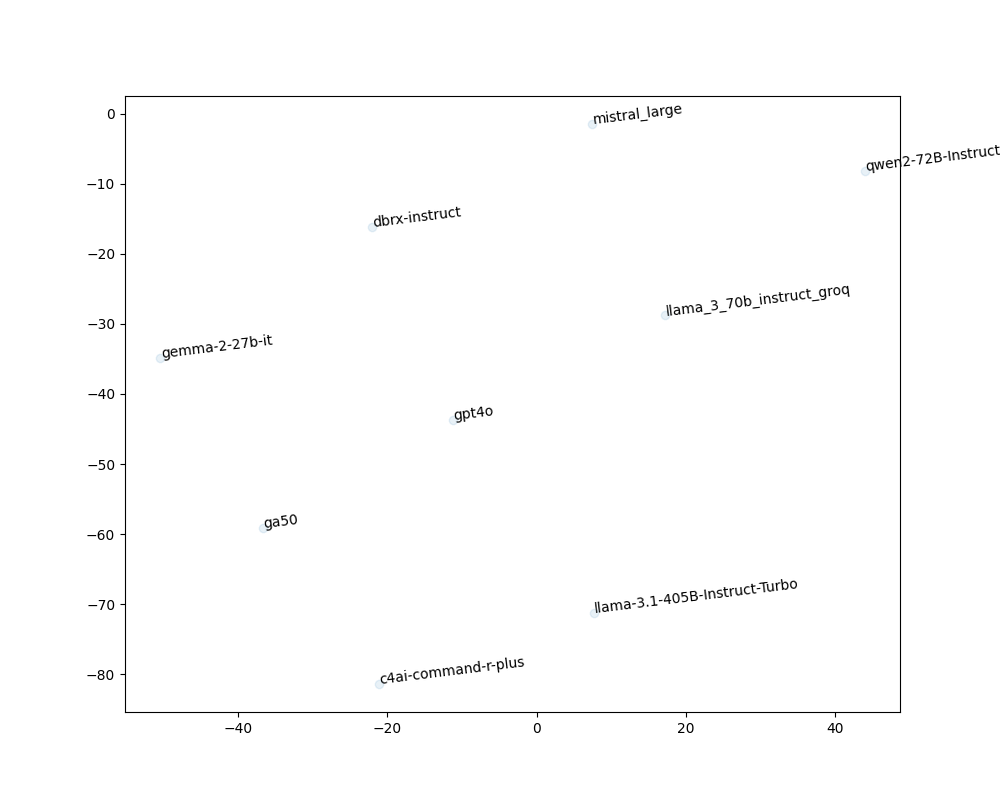}}%
\caption{t-SNE representation of graphs returned by the prompted LLMs.}
\label{a:tsnes}
\end{figure}

Statistics related to Section \ref{s:topologies} appear in Tables \ref{table:stats-lm} and \ref{table:stats-ga50}, as well as the embeddings for these 2 graphs in Figure \ref{a:tsnes}.

\section{Spectral Distances from Ground Truth Graphs}
\label{appendix:spectral}

Since the graph edit distance is intractable, even in practice from graphs with several tens of nodes, other related distances have been proposed, such as the spectral distance \cite{WILSON20082833}, defined as follows:
$$d(G,G')= \sqrt{\Sigma_i (s_i - s'_i)^2},$$ with $s$ the set of eigenvalues $s=\{\lambda_1, \lambda_2, \dots, \lambda_{|V|}\}$, knowing that $\lambda_1 \leq \lambda_2 \leq \dots \leq \lambda_{|V|}$. As recommended in \cite{WILSON20082833}, if graphs are of different sizes, the missing eigenvalues of the smaller are padded with zeros.

We report the spectral distances of LLMs to the KC graph in Table \ref{a:spectral}.

\begin{table}[h!]
\centering
\scriptsize{
\begin{tabular}{|c|c}
\hline
\cellcolor[HTML]{DDDDDD}\textbf{LLM} & \multicolumn{1}{c|}{\cellcolor[HTML]{DDDDDD}\textbf{Spectral distance \cite{WILSON20082833}}} \\ \hline
(reference: KC ground truth)\cellcolor[HTML]{EEEEEE}  &  0 \cellcolor[HTML]{EEEEEE}  \\ \cline{1-1}
mistral-large  &  23.6 \\ \cline{1-1}
dbrx-instruct  &  27.34 \\ \cline{1-1}
gpt3.5  &  28.1 \\ \cline{1-1}
gpt4o  &  28.1 \\ \cline{1-1}
llama-3-70b-instruct-groq  &  32.64 \\ \cline{1-1}
qwen2-72B-Instruct  &  33.41 \\ \cline{1-1}
llama-3.1-70B-Instruct-Turbo  &  35.25 \\ \cline{1-1}
c4ai-command-r-plus  &  36.1 \\ \cline{1-1}
gemini  &  37.96 \\ \cline{1-1}
llama-3.1-405B-Instruct-Turbo  &  37.33 \\ \cline{1-1}
llama-3-70B-Instruct-Lite  &  37.63 \\ \cline{1-1}
llama-3-70B-Instruct-Turbo  &  37.81 \\ \cline{1-1}
llama-3-sonar-large-32k-chat  &  38.19 \\ \cline{1-1}
llama-3.1-8B-Instruct-Turbo  &  38.52 \\ \cline{1-1}
llama-3-sonar-small-32k-chat  &  38.6 \\ \cline{1-1}
llama-3-8b-instruct  &  38.61 \\ \cline{1-1}
phi-3-mini-4k-instruct  &  39.24 \\ \cline{1-1}
gemma-2-27b-it  &  39.32 \\ \cline{1-1}
llama-2-70b-chat-hf  &  39.54 \\ \cline{1-1}
llama-2-13b-chat-hf  &  39.6 \\ \cline{1-1}
\end{tabular}
}
\caption{The spectral distance of the graphs returned by 21 LLMs when prompted for the Zachary's Karate Club graph.}
\label{a:spectral}
\end{table}

Note that as this distance is not the same as the distance between degree sequences from Table \ref{table:stats}, we observe changes in the LLM ranking, such as with mistral-large leading Table \ref{a:spectral}. In this precise case, it appears that its high hallucination in the number of edges returned account less with this spectral metric.

\section{Experiment settings}
\label{appendix: Experiment settings}
\subsection{Prompts for Erdős–Rényi generations}

Here are the two prompts used to generate Erdős–Rényi graphs with LLMs. The \( n \) and \( p \) parameters are \textcolor{blue}{blue-marked} to signify they vary (since we adjust the tested parameters). As an example, we used \( n = \textcolor{blue}{15} \) and \( p = \textcolor{blue}{0.3} \).

\textbf{Direct Prompt:}
\texttt{
\\Give a realization of an Erdős–Rényi graph of parameters n=\textcolor{blue}{15},  p=\textcolor{blue}{0.3}. \\
Use only this format: edges are enclosed in parentheses, integers are separated by a comma, and edges are separated by commas. Do not include any additional explanations or text, only the edges of the graph in the exact specified format.
}

\textbf{CoT Prompt (Chain of Thought):}
\texttt{
\\Give a realization of a Erdős–Rényi graph of parameters n=\textcolor{blue}{15}, p=\textcolor{blue}{0.3}.\\
Think step by step. At the end of your reasoning, please say "final answer: " and use this format for the final answer: edges are enclosed in parentheses, integers are separated by a comma, and edges are separated by commas.
}
\subsection{Sampling parameters}
The temperature used is always mentionned on the figures (varies from 0.2 to 2.0 with a .2 step).
The top-k parameter is set to $50$ and the top-p to $1$ which are default values of the Transformers library (https://huggingface.co/docs/transformers/en/main\_classes/text\_generation).

\subsection{Regex based script}

For the \textit{Direct Prompt}, the script returns the first matching with an edge list of format like \texttt{(1,3) (2,8) (7,2)}. 
For the \textit{CoT Prompt} the script first searches for the occurence of \textit{final answer:} (or similar) and then search for an edge list like above.

\subsection{List of Abbreviations}
\begin{itemize}
    \item LLM: Large Language Model
    \item MCQ: Multiple Choice Question
    \item KC: Karate Club
    \item LM: Les Misérables
    \item GAD: Graph Atlas Distance
    \item CoT: Chain of Thought
    \item GNN: Graph Neural Network
\end{itemize}

\end{document}